# Adapting the Tesseract Open-Source OCR Engine for Tamil and Sinhala Legacy Fonts and Creating a Parallel Corpus for Tamil-Sinhala-English

Charangan Vasantharajan
Dept. of Computer Sci. and Engineering
University of Moratuwa
Colombo, Sri Lanka
charangan.18@cse.mrt.ac.lk

Laksika Tharmalingam
Dept. of Computer Sci. and Engineering
University of Moratuwa
Colombo, Sri Lanka
laksika.19@cse.mrt.ac.lk

Uthayasanker Thayasivam
Dept.of Computer Sci. and Engineering
University of Moratuwa
Colombo, Sri Lanka
rtuthaya@cse.mrt.ac.lk

*Abstract*—Most low-resource languages do not have the necessary resources to create even a substantial monolingual corpus. These languages may often be found in government proceedings but mainly in Portable Document Format (PDF) that contains legacy fonts. Extracting text from these documents to create a monolingual corpus is challenging due to legacy font usage and printer-friendly encoding, which are not optimized for text extraction. Therefore, we propose a simple, automatic, and novel idea that can scale for Tamil, Sinhala, English languages, and many documents along with parallel corpora. Since Tamil and Sinhala are Low-Resource Languages, we improved the performance of Tesseract by employing LSTM-based training on more than 20 legacy fonts to recognize printed characters in these languages. Especially, our model detects code-mixed text, numbers, and special characters from the printed document. It is shown that this approach can reduce the character-level error rate of Tesseract from 6.03 to 2.61 for Tamil (-3.42% relative change) and 7.61 to 4.74 for Sinhala (-2.87% relative change), as well as the word-level error rate from 39.68 to 20.61 for Tamil (-19.07% relative change) and 35.04 to 26.58 for Sinhala (-8.46% relative change) on the test set. Also, our newly created parallel corpus consists of 185.4k, 168.9k, and 181.04k sentences and 2.11M, 2.22M, and 2.33M Words in Tamil, Sinhala, and English respectively. This study shows that fine-tuning Tesseract models on multiple new fonts help to understand the texts and enhances the performance of the OCR. We made newly trained models and the source code for fine-tuning Tesseract, freely available.

*Index Terms*—Tesseract, Printed Character Recognition (PCR), Parallel Corpus,

## I. INTRODUCTION

In the current climate, monolingual corpus for any language is crucial, and with the advent of embedding, the need for the monolingual corpus is increasing [1]. A corpus is a collection of pieces of language text in electronic form, selected according to external criteria to represent, as far as possible, a language or language variety as a source of data for linguistic research. A monolingual corpus is a text corpus that contains only one language. However, we lack such corpora for Low-Resourced Languages (LRL). LRL can be defined as languages that do not have much data or tools available online. Most NLP researchers follow data-driven approaches. Thus, the enhancement of NLP in those languages has been limited so far. A recent study revealed that "the first half-century of research in computational linguistics from circa 1960 up to the present has touched on less than 1% of the world's languages only" [2]. Further, the parallel corpus (corpora that consist of two or more monolingual corpus) would aid research and development in machine translation and language interoperability [3].

Though LRL has not gained much traction in resource building, the need for technologies to process them is growing faster [2]. A larger monolingual corpus is essential for the development of NLP in a specific language. As a first step, we must create such corpora in these languages. It is very common to find the usage of these languages in respective government documents. However, the government documents are primarily Portable Document Format (PDF) with legacy fonts. Besides, in general, these fonts will not be embedded in those PDFs. Even after the standardization of Unicode, the documents in LRL have been mostly created with these legacy fonts. Hence, such text extraction is challenging.

Text extraction from a PDF is only performed if the complete font encoding information is available. After the standardization of Unicode, the text can be extracted from PDFs with Unicode encoding. However, extracting text from a PDF with legacy font requires complete font encoding information. Initially, the discovery of font definitions is needed. This is another challenge in standard text extraction from PDFs. Fonts may be embedded in the PDFs and make discovery easy. If not, we need to search font repositories to find the right fonts to interpret the PDFs. This becomes even more challenging if the fonts used are legacy fonts and are not maintained anymore. For example, the Sri Lankan government's 2017 gazette uses more than 20 Tamil and Sinhala legacy fonts.

In this study, we developed a simple but effective approach that yields high-quality, large-scale trilingual data in Tamil, Sinhala, and English using Deep Learning-based Printed Character Recognition (PCR). For our experiments, we used Tesseract[1] which is an open source text recognition (OCR)

---
[1] https://tesseract-ocr.github.io

Engine. Finally, our approach addresses the text extraction efficiently as well as effectively from the documents which are using legacy fonts.

Our approach distinguishes itself from other approaches in the following ways:
- Using portable government documents to build a document-aligned corpus that helps attain quality exact parallel corpora.
- It is independent of any font usage or embedding.
- Capable of extracting text consisting of all three languages and special characters.

To aid the community of NLP, our contributions will be:
- Deep learning-based models for text extraction from Tamil, Sinhala, and English PDFs/images.
- Document-aligned parallel corpus for Tamil, Sinhala, and English.
- We made our fine-tuned models and source code used for the experiments, publicly available at GitHub[2].

The rest of the sections in the paper are as follows. Section II reviews related experiment works in Corpus creation for low-resourced languages and Tesseract OCR. Section III describes the ground truth generation, model training, and the results with an analysis of the model adaptation process. The fourth section (IV) presents the proposed model. Section V discusses the steps for creating the parallel corpus by using our proposed model and its statistics. Finally, the conclusion is followed by future research directions.

## II. Related Work

Being one of the prominent sub-fields of Computer Science, Natural Language Processing is drastically progressing in the modern era. For the last three decades, it has drawn the attention of most of the world. However, as [2] pointed out, only 1 percent of the languages have been explored reasonably due to the availability of language resources such as corpora in NLP. With the advent of supervised data-demanding approaches like deep learning, these under-resourced languages are side-lined. The importance of a corpus for developing NLP applications for indigenous languages of America, which are also considered LRL, was highlighted in [1]. Importantly developing parallel corpus for low-resourced languages help interoperability and machine translation.

Though developing high-quality and large-sized parallel corpora for many languages is a huge challenge, it is viable for some languages with a web presence, specifically Wikipedia. The general web can be used as a parallel corpus, as explained by [4]. They insisted on creating corpora from various online sources on the web. However, this is not the scenario for many LRLs. Moreover, these parallel corpora are not exact translations. As pointed out by [5], the web cannot be used as a potential corpus for many LRLs because even the web is not consisting of enough resources for LRL, and there are so many other factors deciding the capabilities of the web as a corpus [6] have explained how economic, social, and political factors were playing a vast role in endangering languages by limiting their scope on the web. Thus, we can understand how creating a corpus from the web is a limited option for an LRL and limits its progress in NLP.

In contrast to previous approaches; we focus on using government documents as they are exact translations. However, these documents are mostly portable in legacy fonts. To extract the text from a PDF, we must be aware of the font encodings. Since we mostly do not have the encoding information, traditional PDF tools fail to extract the text. Therefore, many researchers worked on various mechanisms to identify the encodings [7]. Moreover, [8] has proposed a new way for automatic legacy font identification. But still, they did not work out well for PDF text extraction. Therefore, the researchers started to use Optical Character Recognition for text extraction. As per [9], text extraction has four main parts using OCR. They are layout analysis, segmentation, character recognition, and structure recognition. Additionally, [10] highlighted how layout analysis can enhance text extraction precision.

OCR is unconcerned with segmentation and layout analysis. So we propose a layout analysis-based text extraction process on the trilingual government data set, which would produce quality and scalable corpora. Moreover, this effort gains more importance as an approach applicable to several low-resourced languages and the first effort to create a trilingual parallel corpus in Sinhala, Tamil, and English.

## III. Model Adaptation

The Tesseract models are performed well on the text that is generated using widely used fonts of both high-resource and low-resource languages. For high-resource languages, the Tesseract models has been trained on 400000 lines of text spanning about 4500 fonts[3]. In our case, if we consider lower-resource (i.e., Tamil or Sinhala) language models, those are trained on a small number of fonts but on a similar number of text lines as high-resources languages. This worked for problems close to the training data but different in some subtle way, like a particularly unusual font (legacy). Therefore, it is beneficial to have more fonts, as neural networks do not generalize and need to train on the target domain. There are multiple options for training on new fonts: Fine-tune, cut off the top layer (or some arbitrary number of layers) from the network, retrain a new top layer using the new data, and retrain from scratch; we have decided to go with Fine-tune. Fine-tune is a process that takes a model that has already been trained for one given task and then tunes the model to make it perform a downstream task. In this study, we fine-tuned Tamil and Sinhala Tesseract models on the legacy fonts which are frequently used in the Sri Lankan government documents.

### A. Ground Truth Data Generation

Our deep learning-based PCR to extract text from the PDF files depends mainly on how successfully we train the model.

---

[2]https://github.com/aaivu/Tamizhi-Net-OCR

[3]https://github.com/tesseract-ocr/tesseract

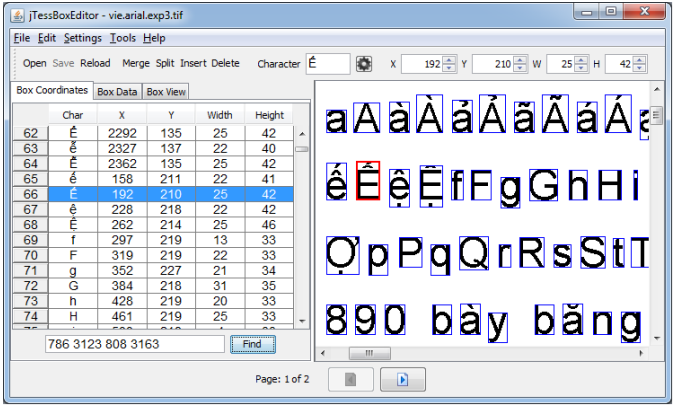

Figure 1: Sample rendering of a TIFF file in jTessBoxEditor. Source image: https://vietocr.sourceforge.net/training.html

Since we are focusing on lower-resource languages, there are no ground truth data with enough image files for every font, letter, and special character to train the model. So, we created the ground truth files for our experiments by using a training text file on the target fonts. For the training text file, we need comparatively large text for each font with enough recurrences of every letter and special characters to train as much as possible and to increase the accuracy and precision. We used the training text file[4] which is provided by tesseract.

After getting the text file, we carefully identified 10 Tamil and 10 Sinhala fonts which are mostly used in Sri Lankan portable documents, and downloaded them from Free Tamil Font[5] website and Sinhala Fonts[6] website. Then, we created the TIFF/Box pair of files for Tamil, and Sinhala using the downloaded fonts. Each font is mapped with a TIFF file that contains 250 pages of images.

From the multi-page TIFF files, we created box files with coordinates specification, and then we rectified misidentified characters, adjusted letter tracking, or spacing between characters to eliminate bounding box overlapping issues using **jTessBoxEditor**[7](Figure 1). Finally, the deep learning model implemented this using **Tesseract**, was trained by using the TIFF/Box pair of files.

Moreover, we have used `tessdata_best`(these are the most accurate trained LSTM models) and `langdata_lstm` (data used for LSTM model training) from Tesseract as our language model and language data.

### B. Model Training

During the training, with base Tesseract, a starter trained data file (tessdata_best[8]) was given for each language and had to be set up in advance. It contains:

- Config file providing control parameters.

---

[4]https://github.com/tesseract-ocr/langdata_lstm/blob/main/tam/tam.training_text
[5]https://www.freetamilfont.com
[6]https://sinhala-fonts.org
[7]https://vietocr.sourceforge.net/training.html
[8]https://github.com/tesseract-ocr/tessdata_best

Table I: The table illustrates the command line flags used during the training. We have finalized the above numbers after conducting several experiments with different values.

| Flag | Value |
|---|---|
| traineddata | Path of the training data file that contains the unicharset, word dawg, punctuation pattern dawg, number dawg |
| model_output | Path of output model files / checkpoints |
| learning_rate | 1e-05 |
| max_iterations | 5000 |
| target_error_rate | 0.001 |
| continue_from | Path to the previous checkpoint from which to continue training. |
| stop_training | convert the training checkpoint to the target model. |
| train_listfile | Filename of a file listing training data files. |
| eval_listfile | Filename of a file listing evaluating data files. |

- Unicharset defining the character set.
- Punctuation pattern dawg, with patterns of punctuation allowed around words.
- Word dawg. The system word-list language model.
- Number dawg, with patterns of numbers that are allowed.

To reach a high accuracy, we want to choose high iterations for training, but it will take too much time. Instead of taking a few minutes to a couple of hours to train, Tesseract 4.1.1 takes nearly two weeks on Nvidia GeForce MX350. Therefore, we decided to train our model for several steps by writing checkpoint files. This allows training to be stopped and continued again later. We periodically wrote checkpoint files at new bests achieved during training. Then, we used the `--stop_training` command line flag to convert any checkpoint to trained data and called `--continue_from` either an existing checkpoint file or from an extracted LSTM model file to modify the network and retrain the remaining. Moreover, Table I summarises `lstmtraining` command-line options.

### C. Experimental setup and Performance evaluation

The common way of measuring the performance of the model is with the accuracy metric, but this does not provide enough granularity to assess OCR performance effectively. In this regard, the error rate is used instead of accuracy to determine how OCR transcribed text and ground truth text differ from each other.

In this analysis, we consider two metrics to evaluate OCR output, namely Character Error Rate (CER) and Word Error Rate (WER).

*1) Character Error Rate (CER):* CER calculation is based on the concept of Levenshtein distance, where we count the minimum number of character-level operations required to transform the ground truth text (aka reference text) into the OCR output.

CER is represented with the following formula.

$$CER = \frac{S + D + I}{N} \quad (1)$$

Where S = Number of Substitutions, D = Number of Deletions, I = Number of Insertions, N = Number of characters in reference text (aka ground truth). The output of this equation represents the percentage of characters in the reference text that was incorrectly predicted in the OCR output. The lower the CER value (with 0 being a perfect score), the better the performance of the OCR model.

Table II: The table shows the evaluation metrics of some trained Tamil fonts. NoC: Number of Characters, RC: Recognized Characters, CER: Character Error Rate, WER: Word Error Rate.

| Font | NoC | Original Tesseract | | | Fine-tuned Tesseract | | |
|---|---|---|---|---|---|---|---|
| | | RC | CER (%) | WER (%) | RC | CER (%) | WER (%) |
| Aabohi | 757 | 757 | 0.19 | 2.67 | 757 | 0.19 | 2.67 |
| AnbeSivam | 762 | 774 | 7.87 | 57.89 | 765 | 2.71 | 31.58 |
| Baamini | 762 | 770 | 7.44 | 56.26 | 762 | 2.42 | 31.58 |
| Eelanadu | 762 | 773 | 4.88 | 43.42 | 763 | 0.58 | 9.21 |
| Kamaas | 762 | 756 | 3.38 | 28.95 | 766 | 0.43 | 9.21 |
| Keeravani | 767 | 764 | 0.68 | 13.16 | 764 | 0.19 | 1.32 |
| Kilavi | 762 | 767 | 0.48 | 9.21 | 763 | 0.14 | 2.63 |
| Klaimakal | 762 | 765 | 0.82 | 14.47 | 766 | 0.48 | 3.95 |
| Tamilweb | 762 | 808 | 20.39 | 88.89 | 772 | 11.13 | 67.90 |
| Nagananthini | 762 | 783 | 14.2 | 82.89 | 785 | 7.83 | 46.05 |
| Mean | | | 6.03 | 39.68 | | 2.61 | 20.61 |

Table III: The table shows the evaluation metrics of some trained Sinhala fonts. NoC: Number of Characters, RC: Recognized Characters, CER: Character Error Rate, WER: Word Error Rate.

| Font | NoC | Original Tesseract | | | Fine-tuned Tesseract | | |
|---|---|---|---|---|---|---|---|
| | | RC | CER (%) | WER (%) | RC | CER (%) | WER (%) |
| Bhasitha | 731 | 701 | 25.97 | 84.62 | 725 | 8.73 | 46.15 |
| BhashitaComplex | 731 | 728 | 5.11 | 27.35 | 731 | 3.94 | 23.08 |
| Bhasitha2Sans | 731 | 726 | 4.68 | 23.93 | 730 | 3.88 | 22.22 |
| Bhasitha Screen | 731 | 726 | 4.79 | 24.79 | 729 | 3.99 | 23.93 |
| Dinaminal Uni Web | 731 | 728 | 5.64 | 29.91 | 731 | 4.52 | 22.22 |
| Hodipotha | 731 | 726 | 6.07 | 35.90 | 729 | 4.10 | 24.79 |
| Malithi Web | 731 | 718 | 6.01 | 34.19 | 726 | 4.74 | 29.91 |
| Noto Sans Sinhala | 731 | 730 | 3.94 | 23.08 | 732 | 3.73 | 21.37 |
| Sarasavi Unicode | 731 | 709 | 9.10 | 38.46 | 728 | 5.64 | 27.35 |
| Warna | 731 | 726 | 4.74 | 28.21 | 732 | 4.10 | 24.79 |
| Mean | | | 7.61 | 35.04 | | 4.74 | 26.58 |

*2) Word Error Rate (WER):* Word Error Rate might be more applicable if it involves the transcription of paragraphs and sentences of words with meaning (e.g., pages of books, and newspapers). The formula for WER is the same as that of CER, but WER operates at the word level instead. It represents the number of word substitutions, deletions, or insertions needed to transform one sentence into another. WER is represented with the following formula.

$$WER = \frac{S_w + D_w + I_w}{N_W} \quad (2)$$

To evaluate, we run the open-source Tesseract OCR model and our fine-tuned model to extract output from several sample images of text. We then utilized the **fastwer**[9] package to calculate CER and WER from the transcribed output and ground truth text (which we labeled manually). The Tables II and III indicate the metrics of Tamil and Sinhala respectively.

*3) Experimental setup:* We prepare test images (every sample image consists of 762 characters and 77 words) from some randomly selected fonts to compare the existing Tesseract model with our trained model according to the above-defined error rates. Table II and III summarise the comparison results.

*4) Performance evaluation:* The quality difference between the existing Tesseract and its fine-tuned model is obvious due to the inability to recognize and render some characters in the Tamil and Sinhala languages. When we extracted the text using the existing model, some characters were missing/misidentified for several fonts as described in Table II and III. This shows the limited capabilities of the existing model when it comes to legacy fonts.

## IV. TAMIZHI-NET OCR

Once we trained our PCR models, we began to extract data from Tamil and Sinhala PDF/Images irrespective of the font. Unlike the traditional approach of using various font encryptions, the accuracy and precision of this method depend only on how we trained our model and processed the input document (Figure 2 illustrates the architecture of our approach). If the input file is PDF, then we will convert it into images. Otherwise, we directly use that image for the next step. For each image in the input, pre-processing the image through some advanced steps by using OpenCV[10] and recognizing the characters slightly increased the accuracy. Note, that we built an independent model for each language and used a hybrid approach that can handle code mix and special characters in a single PDF document

Normally OCR takes image files as the input, but in our case, most government documents are PDFs, so we have developed an algorithm (as shown in Algorithm 1) to handle PDF documents; we use a `filetype` python library to detect file types.

---

**Algorithm 1** Algorithm for Tamizhi-Net OCR Workflow

**Input**: String fileName
**Output**: extracted text file
1: **procedure** TAMIZHI-NET($fileName$)
2:     Initialization: config = '–oem 3 –psm 1'
3:     **if** filetype.guess($fileName$) = 'pdf' **then**
4:         pages = convert_from_pdf($fileName$)
5:         output = []
6:         LOOP Process
7:         **for** $i \leftarrow 0$ to $len(pages)$ **do**
8:             text = ocr_driver(pages[i])
9:             output.append(text)
10:         **return** $joinPages(output)$
11:     **else**
12:         text = ocr_driver($fileName$)
13:         **return** $text$

---

### A. Pre-processing Module

It is no secret that no model is perfect without pre-processing. After the training, we tested our model without any pre-processing, but the model performed poorly when the image had a lot of noise. Other conditions like brightness or skewness of text will also affect the performance of our model. Unfortunately, many input images will contain a plethora of objects and not just clean pre-processed text. Therefore, it becomes imperative to have a good text detection system that can detect text that can be easily extracted. Therefore, we used OpenCV, which is the classic way of doing text detection, and applied various transformations like image Resizing, Blurring, Thresholding, and Contour Detection. (a short description is given below). Figure 3 shows the Original, Grayscale, Threshold, and Contour detected images in that order. We used OpenCV contours detection to detect contours to extract chunks of data. Finally, we apply text recognition to the contours that we got to predict the text.

---
[9]https://pypi.org/project/fastwer/
[10]opencv.org/

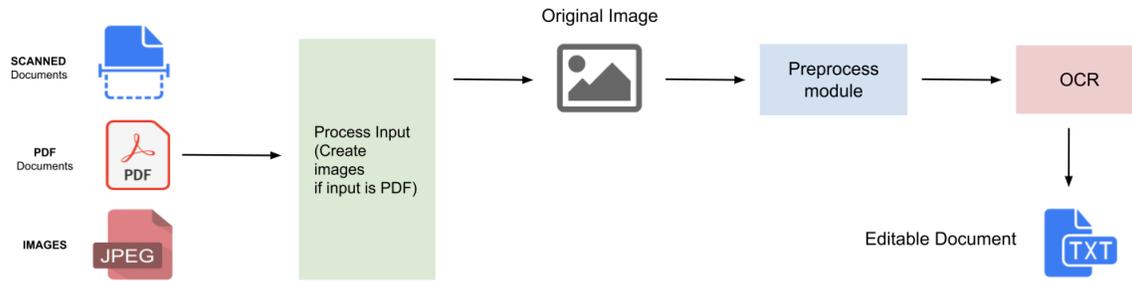

Figure 2: The architectural diagram of our Tamizhi-Net OCR

- Grayscaling
  It is converting a color image to a grayscale. Grayscale is a collection of monochromatic (gray) shades, ranging from pure white on the lightest end to pure black on the opposite end. To get a grayscale image, the color information from each channel is stripped away, leaving only the luminance values. As a result, the image becomes a pattern of light and dark patches devoid of color, essentially a black-and-white image. Figure 3a shows the original image and Figure 3b shows the grayscaled image.
- Resizing
  It is modifying the dimensions of the image, which can be either only width, only height, or both. Also, the aspect ratio of the original image could be preserved in the resized image.
- Image blurring
  It is achieved by convolving the image with the kernel of a low-pass filter. It is effective for noise reduction. It eliminates high-frequency information (such as noise and edges) from the image. Consequently, this process slightly blurs the edges.
- Thresholding
  It is a sort of image segmentation in which the pixels of an image are altered to make it simpler to analyze. In thresholding, a grayscale image is converted to a binary image. White pixels indicate the background, whereas black pixels indicate the foreground. Blurred image is the input to the thresholding operation, which outputs Figure 3c.
- Contour detection
  It is the process of joining all the continuous points along the boundary having the same color intensity. Using contour detection, edges of objects can be detected and localized easily in an image. Figure 3c is the input to the contour detecting operation, which outputs Figure 3d.

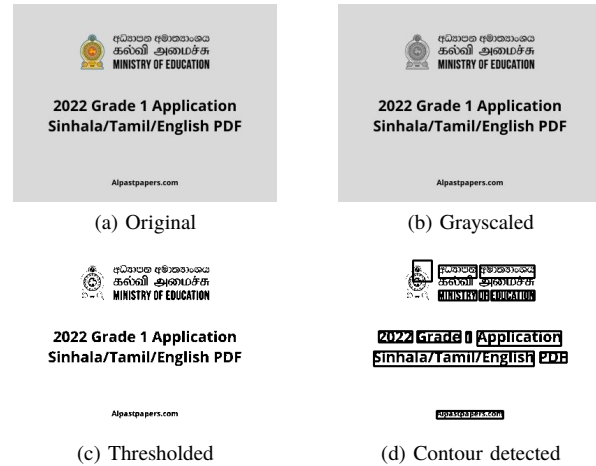

Figure 3: Manipulated images by using Grayscaling, Thresholding, and Contour detecting techniques. When applying the thresholding, the text is only remains in the image, and bounding boxes are created around the text for contour detection.

V. CORPUS CREATION

Tasks like Named Entity Recognition, Translations, and Word embedding creations require a parallel corpus to implement such applications in a multilingual community. But in the case of lower-resource languages like Tamil and Sinhala, such datasets are lacking. So we decided to create a parallel corpus for Tamil-Sinhala-English using our developed OCR.

### A. Documents Collection

For this purpose, we searched for the documents where each document is available in Tamil, Sinhala, and English. Then we found out that the Sri Lankan parliament's presented business papers are available in all three languages. So, we collected parallel documents from the parliament's official websites[11]. The PDF documents are listed according to the date when it happened and have the same title in their language. Then, we downloaded 100 PDF documents for each language and stored them in three different directories with the same name.

### B. Text Extraction using Tamizhi-Net OCR

After collecting the parallel documents, we feed the raw PDFs to our OCR model described in the section IV and get the raw texts. The output file for a PDF is a text file containing its raw text. So we stored the output files in a separate directory and keep the folder structure the same as how the downloaded PDFs. To create a parallel corpus, the collected documents should be monolingual in each language. But in our case, the writers used English abbreviations in Tamil and Sinhala documents. So we try to remove the abbreviations in both languages' documents, but in the meantime, we also need to remove them in the English documents to keep them parallel. It seems simple to remove abbreviations from Tamil and Sinhala documents, whereas it was difficult to remove them from English documents, so we have kept the abbreviations as they are. Therefore the extracted corpora are not going to be a fully monolingual corpus but can be a document-aligned parallel corpus.

### C. Dataset Statistics

Table IV illustrates the final data statistics derived from Tamizhi-Net OCR. Even if we used an equal number of files for each language, the total size of the derived datasets is different in size. Among the three corpora, Tamil has the highest value in terms of document size followed by Sinhala and English respectively. On the contrary, the total number of sentences in Tamil and English corpora is nearly equal and slightly higher than in the Sinhala corpus. According to the Figure 4, the text distribution between the three languages' documents is approximately same, thus indicating our extraction process detected almost all text in all three languages' documents. We also analyze the most common words as shown in Table V by simply ranking the words according to how many times they appear in the body of the text of all documents of every language.

[11]www.parliament.lk/

Table IV: The table describes dataset statistics. NoF: Number of Files, NoS: Number of Sentences, NoW: Number of Words, NoUW: Number of Unique Words, TS: Total Size

| Language | NoF | NoS | NoW | NoUW | TS |
|---|---|---|---|---|---|
| Tamil | 100 | 185.4K | 2.11M | 334.16K | 45.3MB |
| Sinhala | 100 | 168.9K | 2.22M | 407.99K | 35.7MB |
| English | 100 | 181.04K | 2.33M | 372.03K | 20.8MB |

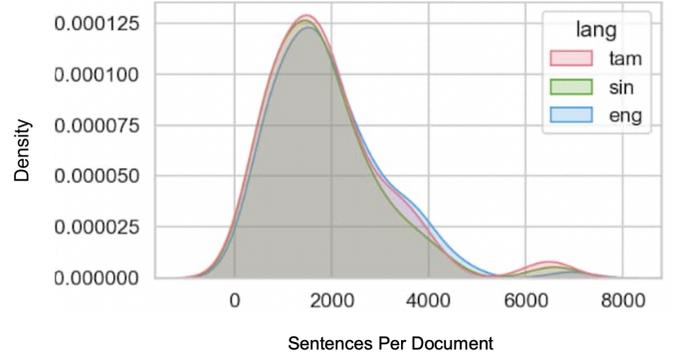

Figure 4: Density plot for the number of sentences per document. Most of the documents in all three languages consist of between 1000 and 2000 sentences.

Table V: The table lists the most frequent words found in the parallel corpus

| | Tamil | | Sinhala | | English | |
|---|---|---|---|---|---|---|
| No | Word | Count | Word | Count | Word | Count |
| 1 | மற்றும் | 29649 | සඳහා | 15058 | Sri | 5065 |
| 2 | ஆம் | 10063 | කිරීම | 10692 | financial | 4618 |
| 3 | தேசிய | 4619 | කරන | 7384 | Lanka | 4051 |
| 4 | இலங்கை | 4585 | මූල්‍ය | 7341 | under | 3896 |
| 5 | அபிவிருத்தி | 3866 | කටයුතු | 5351 | Rs | 3627 |
| 6 | நிதி | 3815 | විසින් | 5127 | National | 3527 |
| 7 | ரூபா | 3610 | අතර | 5027 | Report | 3341 |
| 8 | கணக்காய்வு | 3025 | මුදල් | 4863 | Development | 3118 |
| 9 | அறிக்கை | 2963 | ජාතික | 4432 | Management | 3016 |
| 10 | கீழ் | 2645 | ලංකා | 4302 | Annual | 2986 |

### VI. CONCLUSION

This paper demonstrated that our PCR approach could produce a large-scale quality parallel corpus from government PDFs. Our novel model and pre-processing techniques produced a high-quality output, which makes this created corpus usable for downstream applications. We can include this point in the abstract as well for better attention from readers in low-resource languages such as Tamil and Sinhala with a high-resource language English. The fine-tuned models and the parallel corpus are publicly available at GitHub[12].

### VII. FUTURE WORK

In the above research work, we created a quality large-scale Tamil-Sinhala-English Parallel Corpus Using Deep Learning Based Printed Character Recognition. The next step in our development is to create a spell checker and use it to revise our parallel corpus. It will undoubtedly increase the precision of our dataset. Generally, word usages are different from one domain to another domain. So, without a proper word representation, researchers face difficulties analyzing government documents these days. Therefore, as another step, we are going to create word embeddings for all three languages using the created parallel corpus. So, future researchers can use our word embeddings in their research works.

[12]https://github.com/aaivu/Tamizhi-Net-OCR

## VIII. Acknowledgment

This research was funded by the University of Moratuwa Senate Research Committee SRC/LT/2019/29 & SRC/ST/2019/49. The authors would like to thank Tesseract team for their efforts in making Tesseract open source.